\documentclass[a4paper,conference]{IEEEtran}
\usepackage{graphicx}
\usepackage{mathtools}
\usepackage{xcolor}
\usepackage{float}
\usepackage{multirow}

\begin{document}

\title{Near-Infrared Depth-Independent Image Dehazing using Haar Wavelets}

\author{\IEEEauthorblockN{Sumit Laha, Ankit Sharma, Shengnan Hu and Hassan Foroosh}
\IEEEauthorblockA{Department of Computer Science\\
University of Central Florida\\
Orlando, FL, USA\\
\{sumitlaha,  ankit.sharma285, shengnanhu\}@knights.ucf.edu, hassan.foroosh@ucf.edu}}

\maketitle

\begin{abstract}
We propose a fusion algorithm for haze removal that combines color information from an RGB image and edge information extracted from its corresponding NIR image using Haar wavelets. The proposed algorithm is based on the key observation that NIR edge features are more prominent in the hazy regions of the image than the RGB edge features in those same regions. To combine the color and edge information, we introduce a haze-weight map which proportionately distributes the color and edge information during the fusion process. Because NIR images are, intrinsically, nearly haze-free, our work makes no assumptions like existing works that rely on a scattering model and essentially designing a depth-independent method. This helps in minimizing artifacts and gives a more realistic sense to the restored haze-free image. Extensive experiments show that the proposed algorithm is both qualitatively and quantitatively better on several key metrics when compared to existing state-of-the-art methods.   
\end{abstract}


\IEEEpeerreviewmaketitle

\section{Introduction}
The presence of haze, a common atmospheric distortion, in an outdoor environment results in images with lower contrast, decreased color fidelity, and generally poorer visual quality. This impedes the proper functionality of common computer vision applications, such as obstacle detection and surveillance. Hence, developing image dehazing algorithms is an important step in the computer vision pipeline. The task of image dehazing is to recover a haze-free image from an original, but noisy, image. The widely used model to describe the formulation of a hazy image is: 
\begin{equation}
\label{model_eq}
    \textbf{I} = \textbf{J}(x)\textit{t}(x) + \textbf{A}(1-\textit{t}(x))
\end{equation}

where \textbf{I} and \textbf{J} are the observed intensity and the scene radiance respectively. \textbf{A} is the global atmospheric light and \textit{t} is the medium transmission, which describes the portion of the light that is not scattered and, thus, reaches the camera sensors. Previous works have focused on using this model to develop their solutions by using the intermediate transmission map obtained from it. 

From equation \ref{model_eq}, we can infer that the model characterizes a hazy image as a per-pixel convex combination between the scene radiance and the global atmospheric light. This is an ill-posed problem with at least four unknowns per pixel for an RGB image and inherent ambiguity between haze and scene radiance. To resolve this ambiguity, several works use additional information such as multiple images \cite{narasimhan2003contrast, li2015simultaneous}, scene geometry \cite{kopf2008deep}, or image priors from a single image\cite{he2010single, tan2008visibility}. In this work, we avoid using this physical formulation for modeling the image. Instead, we rely on edge information obtained from the NIR image and fuse it with the color information from the original RGB image to obtain the final dehazed output. The proposed method is depth-independent because it does not produce an intermediate transmission map.

\begin{figure}[t]
\begin{center}
\includegraphics[width=1.0\linewidth]{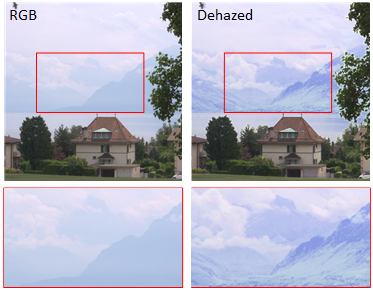}
\caption{Example of a RGB and dehazed image obtained using the proposed approach.\label{fig_eg}}
\end{center}
\end{figure}

In the past decade, several methods have been proposed for image dehazing. Most of the proposed techniques use either single or multiple RGB images to remove haze. The earliest methods utilize multiple images to perform an intensity redistribution via histogram equalization, image luminance, and local image statistics. The presence of additional information, such as having more images in different environments or modalities, can aid in developing better dehazing methods. For example, in \cite{shwartz2006blind} and \cite{schechner2001instant}, two or more images with different degrees of polarization were captured for haze removal, and in \cite{narasimhan2003contrast}, multiple images of the same scene under different weather conditions were used. Recently, single-image-based methods have produced very good results. In \cite{fattal2014dehazing} an image was restored by using a prior wherein colors in local patches lie on a single line in RGB color space. The dark channel prior, introduced in \cite{he2010single}, assumed that there are some low-intensity (i.e. dark) pixels in a color channel within most local patches. This method is efficient, simple and well performed. However, it suffers when there exist large regions of bright surfaces within the scene. In \cite{he2016haze}, a boundary constraint is introduced on the pixel values of the scene transmission when estimating the transmission map.

Unlike the human eye, which is sensitive only to the visible band of the electromagnetic spectrum (380 nm -- 750 nm) \cite{starr2010biology}, a typical camera sensor, which is made of silicon, is sensitive to both the visible and the near-infrared (NIR) bands. To prevent NIR radiation from affecting the color image, most digital cameras place an IR-cut filter in front of the camera sensor. In recent years, the information from NIR imagery has proven very useful in various image processing application \cite{brown2011multi,rufenacht2013automatic,li2007illumination}, especially in image dehazing \cite{feng2013near,son2017near,jang2017colour,dumbgen2018near}.

Rayleigh's scattering law \cite{kim1997contrast} states that the scattered energy $E_s$ is inversely proportional to the fourth power of the radiation wavelength $\lambda$,

\begin{equation}
    E_s \propto  \frac{E_i}{ \lambda ^4} 
\end{equation}

\noindent
where, $E_i$ is the incident radiant energy. This law is applicable as long as the aerosol particles are smaller than one-tenth of the radiation wavelength. Because NIR radiation has a larger wavelength than the visible light, scattering is substantially lower in NIR images than in visible images. Thus, the presence of haze is lower in NIR images as illustrated in Fig.~\ref{fig_eg}. This near haze-free feature of NIR imagery has already been exploited in the past by many researchers \cite{schaul2009color,jang2017colour,dumbgen2018near} for removing haze from RGB image.

In this paper, we propose a fusion algorithm that combines the color information from the luma channel (extracted from the RGB image) and the edge information information from the haze-free NIR image. A simple fusion of both images (i.e. RGB and NIR images) yields poor results. For instance, in a few images the color information in the final dehazed image is not accurate when compared to the original RGB image. In other cases, there can be artifacts present in the final haze-free image. Thus, we introduce a haze-weight map based on the blue channel to proportionately distribute the color and edge information during fusion. This map weights the NIR features (edge information) more and the RGB image (color information) less in hazy regions, and vice versa in the non-hazy regions. Our approach is efficient, unlike the Weighted Least Squares optimization used in several dehazing algorithms presently using NIR imagery \cite{schaul2009color,dumbgen2018near}.

The structure of the rest of this paper is as follows: Section~\ref{sec_literature} reviews the existing literature on the different dehazing algorithms; Section~\ref{sec_method} describes our proposed approach; Section~\ref{sec_results} reports the results of the computational experimentation; and finally, Section~\ref{sec_conc} summarizes our findings and proposes future research directions.

\section{Previous Work}
\label{sec_literature}
We will review two lines of related work: single-image based methods and NIR-based methods.


\subsection{Single-Image based Methods}
Algorithms developed with using only a single hazy image are fast and hence are better suited for real-time applications. In \cite{tan2008visibility}, Tan observe that images taken on clear days have higher contrast than images taken during bad weather. They develop a cost function using MRF with the constraint that the neighboring pixels must have identical airlight values. In practice, this method outputs unrealistic over-enhanced images. In \cite{zhu2015fast}, the authors propose modeling the scene depth of a hazy image and learning the model's parameters in a supervised fashion using a novel color attenuation prior. This method, however, overestimates the transmission in distant regions of the image. Recently, approaches using deep convolutional networks(CNNs) have been proposed with good results. Ren et al. \cite{ren2016single} use two CNNs to restore an image by first feeding the image through a coarse-scale network to obtain the transmission map which is then integrated into one of the intermediate layers of the second CNN (fine-scale network). The fine-scale network then yields the final recovered image. DehazeNet\cite{cai2016dehazenet} uses a CNN based end-to-end system which outputs a transmission map that is then used in subsequent steps to restore the original hazy image.   

\subsection{Near Infrared(NIR) based Methods}
Recently, image dehazing methods have been proposed using NIR information as an additional input with promising results. Schaul et al. \cite{schaul2009color} first proposed using NIR data for image dehazing by transforming visible and NIR images into their respective multiresolution representation using an edge-preserving weighted least squares filter. The restored image given by this method had color-shifting artifacts due to color-enhanced texture regions. In \cite{feng2013near}, the authors introduce a two-step process for image dehazing. The method initially refines the airlight color estimation by taking advantage of differences between the RGB and NIR channels. This is followed by an optimization procedure with NIR constraints to restore the image. It is well known that haze is less severe in longer wavelength bands such as NIR as compared to visible light. The authors in \cite{dumbgen2018near} leverages this knowledge to propose an adaptive hyperspectral algorithm for generating a mask by studying the inconsistency between the intensities of the RGB image and its corresponding NIR image.  

\section{Proposed Method}
\label{sec_method}

\begin{figure}[t]
\begin{center}
\includegraphics[width=1.0\linewidth]{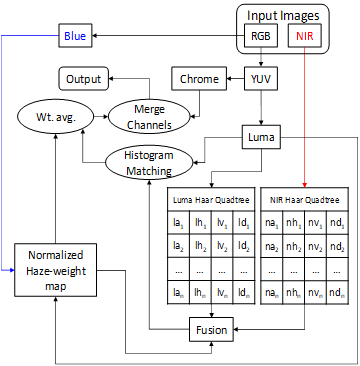}
\caption{Flowchart of the proposed method.\label{fig_flowchart}}
\end{center}
\end{figure}

The main objective of the paper is to recover a clear image by utilizing the edge information from an input NIR image and the color details from the corresponding RGB image. In this paper, we propose a new approach based on Haar wavelets to extract edge information from the NIR image and fuse with the color information of the RGB image. The flowchart of the algorithm is depicted in Fig.~\ref{fig_flowchart}.

\subsection{Luminance channel extraction}

We use the luminance channel of the color image to fuse with the NIR image in our method. Since the human visual system is more sensitive to the luminance \cite{kahu2019review}, we use the luma channel for the subsequent operations. We convert the RGB image to the luminance-chrominance color space. No operation is performed on the chroma channels - they are simply merged with fusion output at the end.

\subsection{Haar wavelet decomposition}

For image fusion, we use an N-level 2-D wavelet transformation with the Haar wavelet \cite{haar1910theorie}. The Haar wavelet function is defined in equation \ref{eq_Haar}.

\begin{equation} \label{eq_Haar}
    \psi(t) =\left\{\begin{matrix}
     1 & 0\leq t< \frac{1}{2}\\ 
    -1 & \frac{1}{2}\leq t< 1\\ 
     0 & \textrm{otherwise}
\end{matrix}\right.
\end{equation}

The function decomposes both the luma and NIR image separately into four components. As depicted in Fig.~\ref{fig_flowchart}, these are coefficients for the approximation, horizontal, vertical, and diagonal components of the decomposition, named as $la$, $lh$, $lv$ and $ld$ respectively for the luma channel and as $na$, $nh$, $nv$ and $nd$ respectively for the NIR channel. We store the values of each level of decomposition for both channels into separate Haar quadtrees. The resolution of each component decreases by half after going through each level of decomposition. After running several experiments, we empirically found out that 2 or 3 levels of decomposition are sufficient for our approach. 

\subsection{Haze-weight map}

\begin{figure}[t]
\begin{center}
\includegraphics[width=1.0\linewidth]{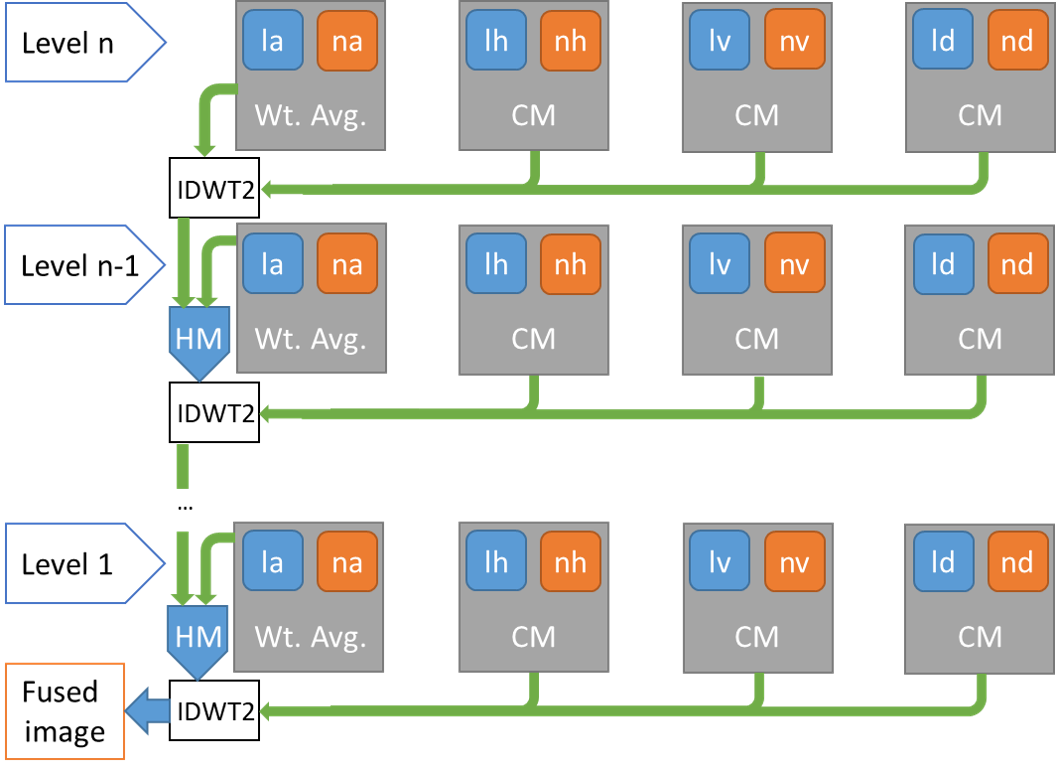}
\caption{Fusion process of Haar coefficients: CM - Choose-max operation, HM - Histogram matching operation, IDWT2 - 2D inverse discrete wavelet transformation.\label{fig_fusion}}
\end{center}
\end{figure}

\begin{figure}[t]
\begin{center}
\includegraphics[width=0.8\linewidth]{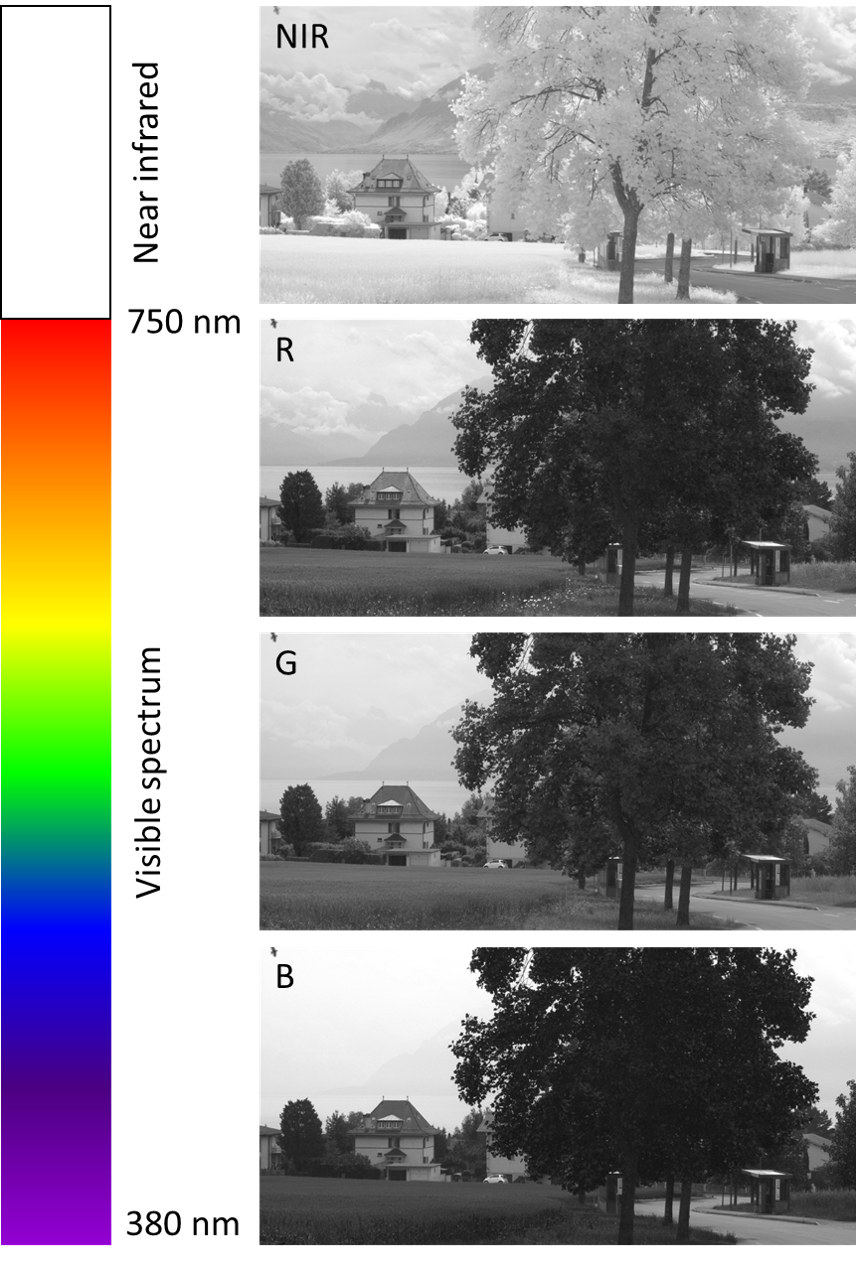}
\caption{Penetration capability of light as observed in different channels \cite{bruno2005crc}.\label{fig_spectrum}}
\end{center}
\end{figure}

The next phase involves the fusion of the luma coefficients with their NIR counterparts as illustrated in Fig.~\ref{fig_fusion}. This is a bottom-up approach, in that we start with the last (coarsest resolution) level of the decomposition and work back towards the original resolution. We design a novel fusion algorithm to combine the respective approximation coefficients from the two images. The key characteristic of our fusion technique is that it selects values from the coefficients in the right proportion, ensuring that the output has the best of both luma and NIR features. The shorter the wavelength of light, the more it scatters due to Rayleigh scattering. Out of the NIR, red, green, and blue channels, the blue channel is most strongly scattered, thus having the lowest penetration power of the four channels. This is illustrated in Fig.~\ref{fig_spectrum}. We utilize this channel to get an estimation of the amount of haze being presented in the scene. This can be achieved by normalizing (0-1 range) the blue channel image to create a haze-weight map ($P$). We use a resized version (corresponding to the dimensions of the coefficients in Haar quadtree) of the map $P$ as weights to compute the weighted average of the luma and NIR approximation coefficients. High intensity in the blue channel means we put more emphasis on the NIR component and vice-versa.

For the fusion of the other three components, we simply compute the co-efficient based activities of each component and combine respective components using the choose-max coefficient combing method \cite{blum2005multi}. Once all the components are fused, we use the inverse Haar wavelet to get an image ($z$) which is double the size of the current level of the Haar quadtree.

We then proceed to the upper level in the Haar quadtree, where the fusion process performed in the previous level are repeated. Because we have the image $z$ from the previous level, an additional operation must be performed. Simply fusing $z$ with an arbitrary component in this level can sometimes create image artifacts. In order to get rid of these artifacts, we match the histogram of the image $z$ with the fused output of the approximation coefficients of this level. The inverse Haar transformation is then applied to the histogram matched image and the other three fused components. The resulting image, once again double in size, updates $z$ for the next level up in the Haar quadtree. Upon fusing the coefficients in the top level in the Haar quadtree, the fusion process ends and the image $z$ is passed on for further processing in the next phase.

\begin{figure}[t!]
\begin{center}
\includegraphics[width=0.8\linewidth]{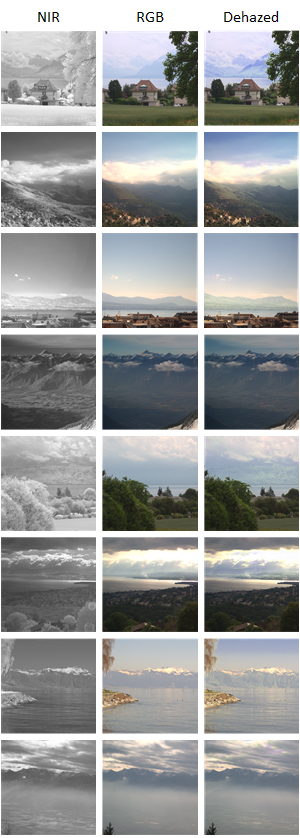}
\caption{Qualitative dehazed results on different images from the RGB-NIR Scene Dataset \cite{brown2011multi}.\label{fig_qual}}
\end{center}
\end{figure}

\subsection{Final image dehazing}

The resulting $z$ now represents the final output of fusion phase. The image $z$ is then histogram matched one again, this time with respect to the original luma image, in order to remove any discrepancies. A final weighted average is then computed between this output and the original luma channel using the haze-weight map $P$ as weights. Finally, the original chroma channels are combined with the weighted-averaged luma channel to produce the final haze-free color image.

\section{Computational Results}
\label{sec_results}

In this section, we compared the performance of our algorithm on 14 images taken from the RGB-NIR Scene Dataset\cite{brown2011multi} with different methods in the literature which represent the state of the art \cite{zhu2015fast,cai2016dehazenet,ren2016single,dumbgen2018near}. We used a small number of images in the experiments due to the availability of limited registered image pairs. Our approach takes into consideration that the RGB-NIR image pair is already registered. We did not use any registration algorithms in our approach. We only ran our method on the image pairs that came registered naturally in the dataset.

\begin{figure}[t]
\begin{center}
\includegraphics[width=1.0\linewidth]{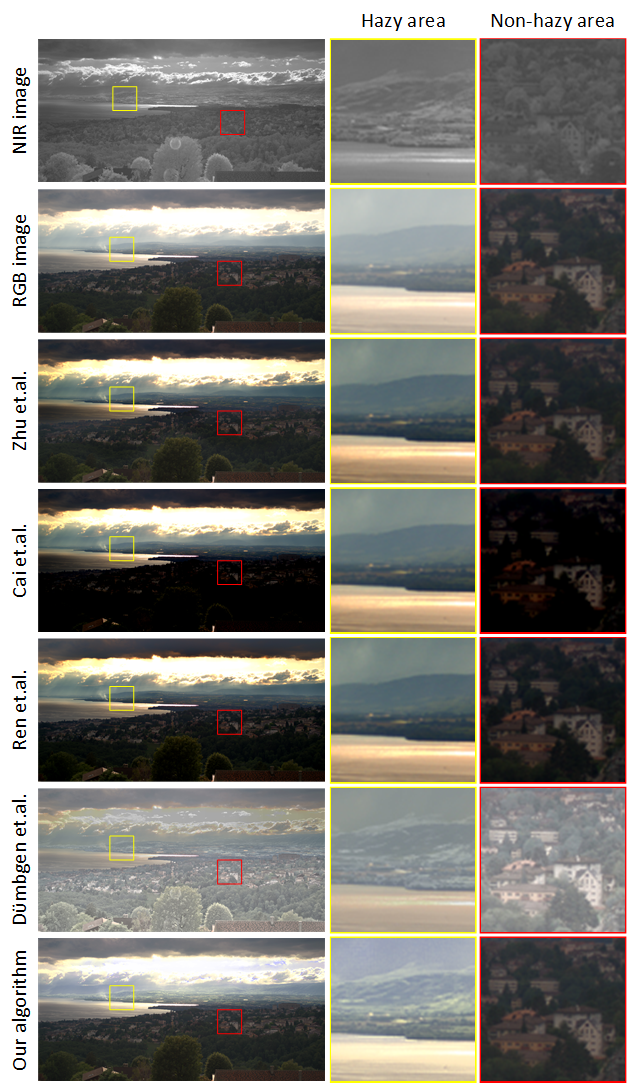}
\caption{Qualitative comparison with state-of-the-art methods\cite{zhu2015fast,cai2016dehazenet,ren2016single,dumbgen2018near}.\label{fig_compare}}
\end{center}
\end{figure}

Since it is very hard to get an accurate evaluation of the dehazed images due to the absence of ground-truth, we provided both qualitative and quantitative analyses of the dehazed images. In the qualitative analysis, we show that the proposed approach removes a considerable amount of haze and/or fog when compared with other methods. In quantitative analysis, we provide some metrics to show that the dehazed images are comparable to the other methods.

\subsection{Qualitative Analysis}

Some of the dehazed images from the proposed approach are illustrated in Fig.~\ref{fig_qual}. A marked improvement in image clarity can be seen in the dehazed images. A visual comparison of our algorithm with the other methods is shown in Fig.~\ref{fig_compare}, wherein we highlight two regions to better show the visual differences between the results. We selected one hazy patch, marked with yellow square, and one non-hazy region, marked with a red square. Our algorithm enhances the edges from the NIR image while preserving the color information from the RGB image in hazy regions without sacrificing the visual quality of the non-hazy regions. Our algorithm maintains almost the same exposure as the original RGB image, whereas the other methods under- or over-expose the non-hazy regions.

\begin{figure*}[t]
\begin{center}
\includegraphics[width=0.8\linewidth]{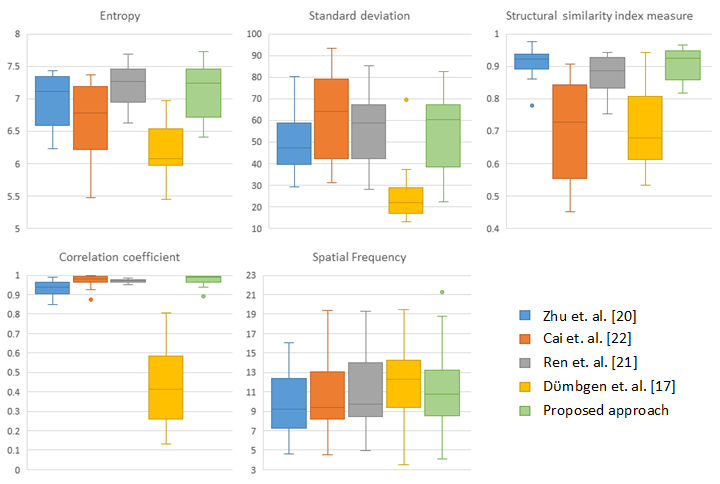}
\caption{Quantitative comparison with state-of-the-art methods\cite{zhu2015fast,cai2016dehazenet,ren2016single,dumbgen2018near} based on the five metrics, i.e., entropy, standard deviation, structural similarity index measure, correlation coefficient and  spatial frequency.\label{fig_quan}}
\end{center}
\end{figure*}

\begin{table*}[t]
\begin{center}
\caption{Quantitative comparison of the proposed method with the state-of-the-art algorithms based on the blind measures of Hautiere et.al.'s method \cite{hautiere2008blind}. The best values are indicated in bold.}
\label{tab_blind}
\begin{tabular}{clrrrrr}
\hline
\multicolumn{1}{l}{} &  & \multicolumn{1}{c}{Zhu et.al.} & \multicolumn{1}{c}{Cai et.al.} & \multicolumn{1}{c}{Ren et.al.} & \multicolumn{1}{c}{D\"{u}mbgen et.al} & \multicolumn{1}{c}{Proposed} \\
\hline
\multirow{3}{*}{country/0000\_rgb} & $e$ & -0.024 & -0.003 & \textbf{0.063} & -0.220 & 0.043 \\
 & $\sigma$ & \textbf{0} & 28.625 & 0.345 & \textbf{0} & \textbf{0} \\
 & $\bar{r}$ & 0.967 & 0.806 & 1.033 & 1.084 & \textbf{1.104} \\
 \hline
\multirow{3}{*}{country/0008\_rgb} & $e$ & \textbf{1.016} & 0.952 & 0.784 & -0.056 & 0.621 \\
 & $\sigma$ & \textbf{0} & 0.002 & 0.027 & \textbf{0} & \textbf{0} \\
 & $\bar{r}$ & 1.462 & 1.310 & 1.456 & 1.556 & \textbf{1.809} \\
\hline
\multirow{3}{*}{country/0021\_rgb} & $e$ & 0.013 & -0.101 & 0.014 & -0.124 & \textbf{0.137} \\
 & $\sigma$ & \textbf{0} & 8.752 & 0.004 & \textbf{0} & \textbf{0} \\
 & $\bar{r}$ & 1.033 & 1.073 & 1.019 & 1.592 & \textbf{1.668} \\
\hline
\multirow{3}{*}{country/0039\_rgb} & $e$ & 0.178 & -0.199 & \textbf{0.370} & -0.030 & 0.200 \\
 & $\sigma$ & \textbf{0} & 38.437 & 1.318 & \textbf{0} & \textbf{0} \\
 & $\bar{r}$ & 1.027 & 0.865 & 1.119 & \textbf{2.581} & 1.270 \\
\hline
\multirow{3}{*}{mountain/0000\_rgb} & $e$ & 0.068 & -0.0134 & \textbf{0.273} & 0.211 & 0.191 \\
 & $\sigma$ & \textbf{0} & 26.381 & 2.639 & \textbf{0} & \textbf{0} \\
 & $\bar{r}$ & 1.017 & 0.790 & 1.048 & \textbf{3.704} & 1.260 \\
\hline
\end{tabular}
\end{center}
\end{table*}

\subsection{Quantitative Analysis}

In recent years, it has been found that no single metric is better than any other for assessing the quality of an image \cite{ma2019infrared}. Consequently, we used multiple commonly used fusion metrics to analyze the images. Apart from the usual image evaluation metrics like entropy, standard deviation, and structural similarity index measure, we also evaluated the images based on correlation coefficient\cite{deshmukh2010image} and spatial frequency\cite{eskicioglu1995image}. Further, we perform blind image assessment on some of the images based on the method used by Hautiere et al. \cite{hautiere2008blind}.

\subsubsection{Fusion Metrics}

The correlation coefficient ($CC$) evaluates the similarity in small size structures between the original RGB image and the dehazed image. It is defined as follows:

\begin{equation} \label{eq_2}
    CC=\frac{\sum\limits_{m}\sum\limits_{n}\left ( I_{mn} - \bar{I} \right )\left ( F_{mn} - \bar{F} \right )}{\sqrt{\left ( \sum\limits_{m}\sum\limits_{n} \left ( I_{mn} - \bar{I} \right )^{2}\right )\left ( \sum\limits_{m}\sum\limits_{n} \left ( F_{mn} - \bar{F} \right )^{2}\right )}}
\end{equation}

\noindent
where, $I$ and $F$ are the original RGB and the dehazed images respectively, and the bar operation denotes a mean over the whole image.

The spatial frequency ($SF$) indicates the overall activity of an image in the spatial domain and is defined as follows:
\begin{equation} \label{eq_3}
    SF=\sqrt{RF^{2}+CF^{2}}
\end{equation}

\noindent
where, $RF$ and $CF$ signifies the row and the column frequencies of the dehazed image ($F$) respectively. They are defined as follows:
\begin{equation} \label{eq_4}
    RF=\sqrt{\frac{1}{mn}\sum_{i=0}^{m-1}\sum_{j=1}^{n-1}\left [ F\left ( i,j \right ) - F\left ( i,j-1 \right )\right ]^{2}}
\end{equation}
\begin{equation} \label{eq_5}
    CF=\sqrt{\frac{1}{mn}\sum_{j=0}^{n-1}\sum_{i=1}^{m-1}\left [ F\left ( i,j \right ) - F\left ( i-1,j \right )\right ]^{2}}
\end{equation}

Although the quality of the images cannot be judged by these metrics, we provided a comparison with the state-of-the-art methods, to show that our images behave in a similar fashion. These comparisons are provided in Fig.~\ref{fig_quan}.

\subsubsection{Blind Image Assessment}

From the method used by Hautiere et al. \cite{hautiere2008blind}, we compare our approach based on the rate of new visible edges ($e$), the quality of contrast restoration ($\bar{r}$), and the number of saturated pixels after restoration ($\sigma$). The higher the values of $e$ and $\bar{r}$, the higher is the image quality. On the contrary, the lower the value of $\sigma$, the better is the quality. A comparison of some of the images is provided in Table~\ref{tab_blind}.

\section{Conclusion}
\label{sec_conc}

In this paper, we address the task of image dehazing by using a pair of RGB and NIR images. We design a fusion algorithm that combines color information from RGB image and edge information from its corresponding NIR image. Existing works suffer from artifacts that produce over-enhanced unrealistic images. To avoid these issues, we devise a method of generating a probability-based haze map which properly weights the color and edge information. Experimental results demonstrate the effectiveness of our proposed method with the final recovered images having better color distribution and revealing more details of the scene. 

\bibliographystyle{IEEEtran}
\bibliography{ref}

\end{document}